\documentclass{article}

\usepackage[preprint]{neurips_2026}
\usepackage{microtype}

\usepackage{pgfplots}
\pgfplotsset{compat=1.18}
\usepgfplotslibrary{fillbetween}
\usetikzlibrary{intersections}
\usetikzlibrary{positioning}
\usepgfplotslibrary{statistics}
\usetikzlibrary{matrix}
\usepgfplotslibrary{groupplots}
\usetikzlibrary{plotmarks}
\usetikzlibrary{shapes.misc}

\usepackage{wrapfig}

\usepackage{amsmath}
\usepackage{amssymb}
\usepackage{siunitx}
\usepackage{bbm}

\usepackage{enumitem}

\usepackage[scaled=0.825]{beramono}
\usepackage[T1]{fontenc}

\usepackage{xcolor}
\usepackage[]{hyperref}
\hypersetup{
  colorlinks,
  linkcolor={red!40!black},
  citecolor={red!40!black},
  urlcolor={red!40!black}
}
\usepackage[nameinlink,capitalise,noabbrev]{cleveref}
\crefname{lstlisting}{listing}{listings}
\Crefname{lstlisting}{Listing}{Listings}

\usepackage{listings}
\usepackage[linesnumbered,ruled,vlined]{algorithm2e}
\crefname{algocf}{alg.}{algs.}
\Crefname{algocf}{Algorithm}{Algorithms}

\usepackage{caption}
\usepackage{subcaption}
\usepackage{placeins}

\usepackage{adjustbox}

\usepackage{tabularx, booktabs}
\usepackage{makecell}
\usepackage{multirow}

\usepackage{ifthen}
\usepackage[inkscapearea=page]{svg}

\usepackage{array}

\lstdefinestyle{sexp}{
    language=Lisp,
    showstringspaces=false,
    basicstyle=\color{codeDarkColor}\ttfamily,
    keywordstyle=\color{codeColor}\ttfamily,
    stringstyle=\color{red}\ttfamily,
    identifierstyle=\color{codeColor},
    mathescape=true
}

\definecolor{poscolor}{RGB}{128, 180, 230}
\definecolor{negcolor}{RGB}{252, 90, 40}
\definecolor{neucolor}{RGB}{1, 74, 26}
\definecolor{color3}{RGB}{230, 128, 225}
\definecolor{color4}{RGB}{152, 190, 87}
\definecolor{color5}{RGB}{11, 49, 66}
\definecolor{gridcolor}{RGB}{85, 85, 85}

\usepackage{pifont} % used to get \ding command
\usepackage{xspace}

\usepackage{enumitem}

\newcolumntype{R}[1]{>{\raggedleft\arraybackslash}p{#1}}

\definecolor{myred}{HTML}{bf616a}

\title{When a Zero-Shooter Cheats: \\Improving Age Estimation via Activation Steering}

\author{%
  Erik Imgrund\\
  BIFOLD \& TU Berlin\\
  \And
  Pia Hanfeld\\
  BIFOLD \& TU Berlin\\
  \And
  Klim Kireev\\
  BIFOLD \& TU Berlin\\
  \And
  Konrad Rieck\\
  BIFOLD \& TU Berlin\\
}

\begin{document}

\maketitle

\begin{figure}[htb]
    \centering
    \input{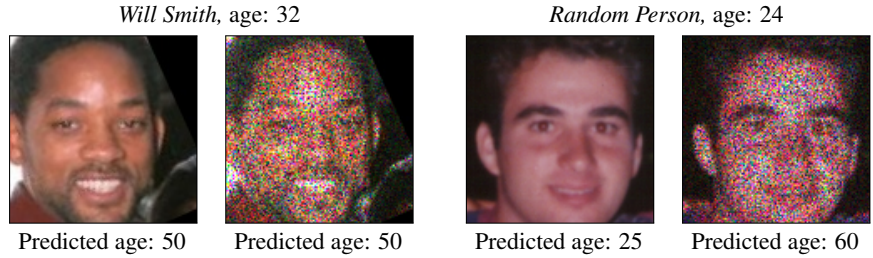}
    \caption{Example of the identity shortcut. The age of Will Smith is consistently mispredicted by Gemma~3 to be close to his age at the knowledge cutoff. The effect persists even in the presence of severe noise, which causes deviations for random people that the model does not recognize. The random person is sampled from the FG-Net dataset, the image of Will Smith from AgeDB.}
    \label{fig:teaser}
\end{figure}

\begin{abstract}
Different age-related regulations have been proposed to protect minors from harmful content and interactions online.
Automated age estimation is central to enforcing such regulations, and vision-language models (VLMs) achieve state-of-the-art performance on this task.
However, we find that the zero-shot nature of VLM-based age estimation produces an unexpected side effect we call the \emph{identity shortcut}: 
Instead of estimating age from visual features, VLMs tend to identify the depicted person and infer their age from memorized knowledge.
This phenomenon leads to substantially incorrect predictions when non-celebrities are misidentified as celebrities. 
It also produces deceptively high robustness to noise and adversarial perturbations on celebrity images, which dominate popular benchmarks.
To mitigate this, we propose an \emph{activation steering} method that suppresses the shortcut by intervening on the hidden states of the VLM. 
This method improves age estimation accuracy for both memorized and unseen identities, reducing mean absolute error by up to 25\% across popular benchmarks.
\end{abstract}
\section{Introduction}

Age estimation from facial images has quietly become safety-critical infrastructure. New legislation in the EU and Australia~\citep{online_safety_amendment_act_2024, europeancommission2025guidelines} require platforms to gate access for minors, and at the scale of deployment, such gating needs to be performed by automated age estimators~\citep{hanaoka2024face,accs2021technical}\footnote{Non-exhaustive list of companies providing certified age verification: \href{https://www.yoti.com/}{Yoti}, \href{https://roc.ai/}{ROC AI}, \href{https://www.scytales.com/}{Scytáles}, \href{https://sumsub.com/}{Sumsub}.}.
%
%Governments worldwide are currently introducing legislation that restricts minors' access to online services and content~\citep{online_safety_amendment_act_2024, europeancommission2025guidelines}. Implementation of these measures at scale often implies automatic age estimation systems~\citep{hanaoka2024face,accs2021technical}\footnote{Non-exhaustive list of companies providing certified age verification: \href{https://www.yoti.com/}{Yoti}, \href{https://roc.ai/}{ROC AI}, \href{https://www.scytales.com/}{Scytáles}, \href{https://sumsub.com/}{Sumsub}}. 
%
Vision-language models (VLMs) have recently surpassed domain-specific image classifiers for this task, demonstrating vastly improved capabilities over the traditional fully-supervised models. However, due to the lack of internal transparency, this improved performance may come at the cost of unexpected behavior.

An example of such counterintuitive behavior is illustrated in \Cref{fig:teaser}. Rather than estimating the age of the depicted person from visual features, the VLM first identifies the subject in the photo and reports their current age from memory, producing a strikingly incorrect prediction. Moreover, unlike any traditional detection method, the prediction does not change even in the presence of high-amplitude noise, implying a persistent underlying bias.

We call this phenomenon the \emph{identity shortcut}: instead of inferring the queried attribute from visual evidence, the VLM first attempts to recognize the subject and reports an attribute associated with it from memory.
This shortcut has severe consequences: 
First, it causes prediction errors on celebrity images, where the model returns an age tied to the memorized identity.
Second, it degrades performance on non-celebrities who visually resemble known figures, as the model anchors its prediction on the closest identity. 
Third, and most insidiously, it manufactures \emph{deceptive robustness}. Models appear exceptionally resilient to perturbations on popular benchmarks, but only because those benchmarks are dominated by celebrity images.

In this paper, we show that this phenomenon is not limited to a specific model or dataset, but instead is a general issue across VLMs. 
To mitigate it, we introduce an inference-time activation steering framework that identifies the hidden-state directions responsible for identity-based retrieval and suppresses them, forcing the model to condition its prediction on visual evidence rather than on a recalled identity. 
Extensive evaluations across state-of-the-art VLMs demonstrate that our intervention not only eliminates the asymmetric robustness between memorized and unseen identities but also reduces mean absolute error by 10.8\% on average (up to 25.3\%) across existing benchmarks.

%Further study of this phenomenon reveals that VLMs silently take shortcuts on some samples. Instead of performing age estimation from the picture, they recognize the depicted identity and then respond with the age of the identified individual. \Cref{fig:teaser} shows this phenomenon and the two resulting issues: older images of celebrities result in predictions of their current age instead of the age in the depicted image and do so with exceptional robustness to corruptions. We show that this can result in woefully incorrect predictions when a person is misidentified as a celebrity. Motivated by those findings, we perform an intervention in the hidden states of the VLM to force disallow this shortcut, resulting in XX\% improved performance and symmetric robustness across all samples.

In summary, we make the following contributions:
\begin{itemize}[leftmargin=*]
    \item \textbf{Identity shortcut:} We introduce the phenomenon of the \emph{identity shortcut}, where VLMs prioritize identity retrieval over visual feature analysis. This effect is responsible for systematic performance degradation and can inflate benchmark results.
    
    \item \textbf{Deceptive Robustness:} We show that zero-shot VLMs exhibit fundamentally different behavior under perturbations than traditional methods, as their inference can be guided by retrieval rather than visual analysis, leading to misleading conclusions about their robustness.

    \item \textbf{Inference-Time Mitigation:} We propose an intervention that steers the model's activations to suppress the identity shortcut. Our method improves accuracy and restores expected robustness, ensuring that age estimates are grounded in visual evidence rather than memorized associations.
\end{itemize}

\section{Preliminaries}
To contextualize our discovery of the identity shortcut, we first review prior work on age estimation and the robustness of VLMs. We then describe the datasets and models that constitute the foundation of our empirical evaluation.

\subsection{Related Work}
Our work sits at the intersection of three active directions in research: methods for automated age estimation from facial images, the robustness of VLMs to input perturbations, and the use of VLMs as zero-shot predictors for image understanding tasks.

\paragraph{Age Estimation.}
Age estimation is an important image understanding task with many applications including enforcing age-based restrictions~\citep{hanaoka2024face}, data mining~\citep{web_ni_2009} or removing underage pictures from image generation datasets~\citep{cretu2025evaluating, gaul_underage_2025}.
% Until the emergence of Vision Transformer~(ViT) models~\citep{vaswani2017attention}, Convolutional Neural Networks~(CNNs) have been the state of the art for classification and regression tasks in computer vision and, subsequently, for age estimation based on images~\citep{roopak_comparison_2023}.
\citet{mean-variance} use convolutional neural networks (CNNs) to perform age estimation based on a cropped image of a person's face. \citet{mivolo_2023} achieve better performance using a vision transformer and combine facial and body images. This approach is further extended by \citet{mivolo2024}, achieving the best performance of any specialized model for age estimation. More recently, \citet{ren_out_2026} show that VLMs achieve even lower error despite lacking any task-specific training. In a related line of research, \citet{szu-tu2025exposing} show that VLMs achieve high performance in estimating the age of buildings through memorization of famous landmarks. We extend this work by quantifying how identity-based biases degrade VLM age estimation and propose a mitigation strategy that restores general performance.

\paragraph{Robustness of VLMs.}
VLMs have been utilized widely for multimodal input analysis across multiple domains~\citep{Li_2025_CVPR}. While most research uses benchmarks that evaluate performance under near-ideal scenarios~\citep{yin_lamm_2023, li_seed-bench_2024, liu_mmbench_2024, fu_mme_2025}, real-world deployments demand accuracy even on degraded inputs, such as blurry or noisy uploads~\citep{ye_rotbench_2024}. Following this work, \citet{usama_analysing_2025} show that VLMs are not robust towards common noise corruptions. Regarding adversarial robustness, \citet{wang_double_2025} demonstrate that adversarial training can restore robust performance to almost clean performance, while \citet{cui_robustness_2024} show that appending LLM-generated descriptions to prompts enhances resistance to white-box attacks. Meanwhile, \citet{Latif_2026_WACV} improve robustness against visual and textual corruption functions by preprocessing the input to the VLM through expert generative denoising models. 

%\citet{cui_robustness_2024} observe that LLM-generated, single-sentence descriptions for a subset of classes from the evaluation dataset added to the prompt can enhance resistance to white-box attacks. Alternatively, \citet{Latif_2026_WACV} improve robustness against visual and textual corruption functions by preprocessing the input to the VLM through expert generative denoising models. However, existing work focuses primarily on input corruptions and assumes a homogeneous data distribution.
However, existing work focuses primarily on input corruptions, implicitly assuming a homogeneous data distribution. We extend this line of work by first investigating the robustness in a regression task, namely age estimation, and comparing robustness properties across different data distributions. We also demonstrate that the robustness of VLMs is dependent on the data distribution, revealing that current state-of-the-art benchmarks provide a false sense of security.
%While clean performance of learned classifiers heavily depends on similarity between the training distribution and test distribution, \citet{naseer2019} show that robustness against targeted perturbations is independent of the underlying distribution, i.e. adversarial perturbations are transferable to previously unseen data distributions. \citet{Hendrycks2021} analyze methods to improve robustness during training of models to prevent performance loss under distribution shifts at inference time. 

%Although many of these concepts have been known for over a decade, there is currently no known method for hardening deep-learning based models against adversarial examples except for adversarial training~\citep{Carlini2017}. To quantify adversarial robustness,~\citet{li2024oodrobustbench} introduce a benchmark for adversarial robustness under distribution shift. The authors observed a tremendous decline in robustness against adversarial examples given out-of-distribution data.
% Based on the findings by \citet{elhage2022superposition}, \citet{gorton_adversarial_2025} argue that adversarially trained models are less prone to adversarial examples. 
% Furthermore,~\citet{salman_adversarially_2020} conclude, that adversarially trained models transfer better to previously unseen image datasets. 
%We extend this research to VLMs and through this manner call the generality of the results of the work introduced to increase robustness into question. 
%We extend this research to VLMs, challenging the generality of existing robustness improvements. 
\paragraph{Zero-shot Reasoning Capabilities.} 
\citet{Nagar2024} raise the question of whether state-of-the-art visual question answering (VQA) benchmarks, developed with CNNs and ViTs in mind, allow meaningful judgement of the actual visual reasoning capabilities of VLMs. 
They design a benchmark on new synthetic data to asses if the zero-shot performance of the models relies on reasoning instead of memorized knowledge. Overall, all evaluated models perform significantly worse compared to a human test group in this study.

\citet{pombal2025} introduce an automatic benchmark for VLMs on a validation set of a public VQA dataset. One VLM is tasked to generate a new question-answer pair from an input image which is then judged by a second model. Their approach automates the generation of VQA tasks for benchmarking from existing, real-world datasets. However, their approach does not explicitly analyze the visual reasoning capabilities of the models. \citet{rizwan_exploring_2025} analyze the capabilities of several VLMs to classify potentially offensive memes. They observe that most models they analyze are rigid about their initial classification even under partial occlusion. 

Overall, extensive work analyzing VLM zero-shot performance remains scarce, with only a few publications explicitly distinguishing between world knowledge and visual reasoning capabilities.
Building on these concerns, we specifically investigate how identity-based memorization compromises zero-shot age estimation. %By comparing model performance under corruptions for celebrities versus non-celebrities, we demonstrate that VLMs frequently bypass visual analysis to rely on the identity shortcut, leading to asymmetric robustness and misleading benchmark scores.

\subsection{Experimental Setup.}
\label{sec:intro:setup}
In this work, we follow the standard evaluation procedure established in the age estimation literature, relying on publicly available datasets and models.

\paragraph{Datasets.} In particular, we use the following three public datasets and one self-collected dataset throughout our study.
\begin{itemize}[leftmargin=20pt]
    \item \textbf{CelebA}~\citep{celebA}, a widely-used dataset containing over 200,000 images of over 10,000 celebrities annotated with 40 binary attributes,
    \item \textbf{AgeDB}~\citep{moschoglou2017agedb}, consisting of over 16,000 images captured under uncontrolled, real-world conditions of 570 celebrities labeled with age and gender,
    \item \textbf{FG-Net}~\citep{Lanitis2002}, composed of 1,002 images of 82 non-celebrities captured at different stages of their lives,
    \item \textbf{VideoScrape}, a dataset consisting of 200 images that we manually scrape from a video platform after the knowledge cutoff of the considered VLMs. We describe it in Appendix~\ref{sec:appendix:datasets}. 
%    collected off a public video platform with upload dates 
\end{itemize}
%VideoScrape consists of images that we manually scraped from videos collected off a public video platform with upload dates after the latest known knowledge cutoff date of our evaluation VLMs.

In order to evaluate our method on roughly equal distributions regarding age and gender, we first gather demographics from our VideoScrape dataset and randomly sample as many images from AgeDB and FG-Net. The demographics for our evaluation sets are reported in \cref{tab:demo_hover} in Appendix~\ref{sec:appendix:datasets}.

\paragraph{Models.} We study different age estimators: six zero-shot vision-language models prompted for age estimation \citep{llava15,qwenvl25,gemma3,qwen3.5,gemma4_modelcard_2026}, and two fully-supervised methods: the state-of-the-art vision transformer MiVOLO~\citep{mivolo2024} and a ResNet-18 trained for age estimation by \citet{mean-variance}. We summarize the information about evaluated methods in \cref{tab:models}.

\begin{table}[tbh]
    \centering
    \caption{Characteristics of the age estimation methods used in our study. The type is either fully-supervised (FS) or zero-shot (ZS). We did not include Gemini 3 Flash in the table, as the details are mostly proprietary.}
    \label{tab:models}
    \vspace*{2pt}
\small
\setlength{\tabcolsep}{5pt}
\renewcommand{\arraystretch}{1.15}
\begin{tabular}{l rr rrrrr}
    \toprule
    & \multicolumn{2}{c}{\textbf{Task-specific}} & \multicolumn{5}{c}{\textbf{Vision-language models}} \\
    \cmidrule(lr){2-3} \cmidrule(lr){4-8}
    & CNN & MiVOLO & LLaVA~1.5 & Qwen-VL~2.5 & Gemma~3 & Qwen~3.5 & Gemma~4 \\
    \midrule
    Parameters   & 12\,M & 27\,M & 7\,B  & 7\,B  & 4\,B  & 9\,B  & 31\,B \\
    Release      & 2018  & 2023  & 2023  & 01/25 & 03/25 & 02/26 & 04/26 \\
    Setting      & FS    & FS    & ZS    & ZS    & ZS    & ZS    & ZS    \\
    \bottomrule
\end{tabular}

\end{table}

\section{Identity Shortcut}

Age estimation systems are assumed to operate on visual features of the input image (face, posture, context, etc.). However, since VLMs are pretrained on vast corpora of data containing factual information about people, they can take a \textit{shortcut} by identifying the depicted person and then performing factual inference from their memorized world knowledge. While one might assume that this mechanism aids performance by retrieving the correct age for identified individuals, it fundamentally deviates from the intended task of visual age estimation. Specifically, returning a person's current age based on the model's knowledge cutoff introduces systematic temporal errors.

Even worse, we find that VLMs frequently attempt to assign identities to non-public figures as well, leading to hallucinated identifications and consequently erroneous age predictions. In this section, we analyze the presence and impact of this \emph{identity shortcut}.

\subsection{Definitions.}
Given an age estimator \(f\), we define \(f(x \mid \mathrm{k})\) as its output on an input image \(x\) drawn from an image distribution \(\mathbb{X}\), with knowledge of the depicted person.
Similarly, \(f(x\mid \neg \mathrm{k})\) denotes that a person on an image \(x\) is unknown to the estimator. \(\mathbb{K}_f\) in our notation is the distribution of images containing people known to the estimator. 
We then define the expected impact \(\delta_k\) of the identity shortcut for an age estimator \(f\) as \begin{equation}
    \delta_k = \mathop{\mathbb{E}}_{x\sim \mathbb{X}} \, | f(x \mid k) - f(x \mid \neg k) |\,.
\end{equation}

\newcommand{\surr}{{g}}

Unfortunately, the direct computation of this term is infeasible, as it would require constructing a model that memorizes exactly the same set of people except for the person shown in the image \(x\). Consequently, we would need to either modify \(x\) so that it is no longer identified correctly, or modify the model itself. Neither option is practical. Instead, we introduce a surrogate estimator \(\surr(\cdot\mid\neg k)\) that is guaranteed not to have specific knowledge of any individual and compute 
\begin{equation}
    \mathop{\mathbb{E}}_{x\sim \mathbb{K}_f} \, |f(x\mid k) - \surr(x\mid \neg k)| \approx \delta_k + E_{f,\surr},
    \label{eq:approxImpact}
\end{equation}
which is accurate up to a modeling error \(E_{f,\surr}\) based on the mismatch between the surrogate and actual estimator. Then, we approximate this term on the distribution of images with unknown identities via 
\begin{equation}
    E_{f,\surr} = \mathop{\mathbb{E}}_{x\sim \mathbb{X} \setminus \mathbb{K}_f} \, |f(x\mid \neg k) - \surr(x\mid \neg k)|\;.
\end{equation}

\subsection{Impact on Known Identities}
\label{sec:meth:celeb}
In order to quantify the effect of memorized identity knowledge on the prediction of VLMs, we 
split the CelebA~\citep{celebA} dataset into two parts, the first one consists of randomly sampled images, and the second one of images where a VLM correctly identifies a celebrity. We measure the difference between the VLMs' age predictions and a baseline age estimator (MiVOLO~\citep{mivolo2024}) that was not trained with knowledge of any identities. The choice of MiVOLO as the baseline estimator is dictated by the following: MiVOLO was trained on manually annotated samples from the Open Images dataset and production data of their age estimation service and both of them are unlikely to have celebrity images. The Open Images dataset contains permissively licensed images from Flickr filtered for images appearing on any other site and therefore have a low probability of including any celebrities.
% The production dataset is likely gathered in Russia (since the model is released to the open-source by a Russian company), and therefore, is also unlikely to contain western celebrities present in CelebA. This absence of CelebA celebrities in the MiVOLO training set makes it a good choice for the baseline estimator.
The resulting modeling error and identity shortcut impact $\delta_k$ are reported in \cref{tab:disagreement}.

\begin{table}[htbp]
    \centering
    \caption{Impact of the identity shortcut \(\delta_k\) estimated using \cref{eq:approxImpact} and the modeling error \(E_{f,\surr}\). All shortcut impacts higher than 1-sigma are shown in bold.}
    \label{tab:disagreement}
    \vspace{2pt}
\begin{tabular}{l ccc}
    \toprule
    Model & \(\delta_k + E_{f,\surr}\) & \(E_{f,\surr}\) & \(\delta_k\) \\
    \midrule
    LLaVa 1.5      & 3.71 $\pm$ 0.20 & 3.46 $\pm$ 0.17 & \phantom{-}\textbf{0.25}  \\
    QwenVL 2.5     & 4.57 $\pm$ 0.24 & 4.16 $\pm$ 0.21 & \phantom{-}\textbf{0.41}  \\
    Gemma 3        & 5.55 $\pm$ 0.25 & 4.39 $\pm$ 0.21 & \phantom{-}\textbf{1.16}  \\
    Qwen 3.5       & 7.82 $\pm$ 0.37 & 7.89 $\pm$ 0.32 & -0.07 \\
    Gemma 4        & 3.32 $\pm$ 0.15 & 3.05 $\pm$ 0.14 & \phantom{-}\textbf{0.27}  \\
    Gemini 3 Flash & 4.48 $\pm$ 0.37 & 3.03 $\pm$ 0.25 & \phantom{-}\textbf{1.45}  \\
    \bottomrule
\end{tabular}
\end{table}

It is immediately apparent that the identity shortcut has a measurable impact on most VLMs. The effect does not show any trend along model size or recency and is strongest for \textit{Gemma 3} and \textit{Gemini 3 Flash}, where its expected impact on the model's prediction is greater than one year. While the outlying behavior of \textit{Qwen 3.5} looks counter-intuitive, we explain it in the following way: \textit{Qwen 3.5} \emph{always} takes the identity shortcut for images in CelebA, even when not correctly identifying them. Indeed, its disagreement with MiVOLO is the highest on both sets, and as we show in Section \ref{sec:meth:steering}, it benefits from the identity shortcut removal the most.

\subsection{Robustness across Distributions}\label{sec:robustness}
Besides the performance impact, we also discover that the identity shortcut has an asymmetric effect on model robustness. We estimate the robustness by applying a set of common corruptions~\(\mathcal{C}\) introduced by~\citep{commonCorruptions} with different noise levels and measure the change in the age estimate. More specifically, we compute the mean deviation across all corruptions. We normalize the deviations of each corruption by the maximum deviation across all data points of all datasets to weigh each corruption inversely proportional to its severity. More formally, we compute \begin{equation}
    \mathop{\mathbb{E}}_{c\in\mathcal{C}, x\in\mathbb{X}} \frac{|f(x) - f(c(x))|}{\max_{x'\in\mathbb{X}}|f(x') - f(c(x'))|}.
\end{equation}

In order to distinguish natural asymmetry on different datasets from the deceptive robustness caused by the identity shortcut, we perform this experiment on multiple datasets (CelebA-K, AgeDB, FG-Net and VideoScrape).

\begin{figure}[tbh]
    \centering
    \input{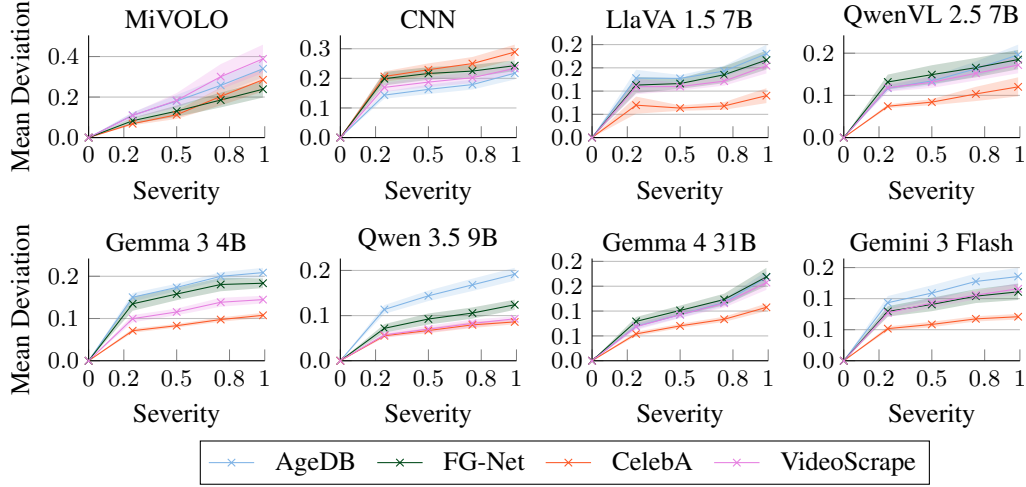}
    \caption{Robustness to common corruptions of different age estimation models. The deviation from the base prediction is averaged over all types of corruption, normalized by the maximum deviation for each corruption. The shaded area shows the standard error of the mean across all corruptions. Vision language models show asymmetrical robustness depending on the data distribution and demonstrate distinctively high robustness on the CelebA dataset.}
    \label{fig:common_corruptions}
\end{figure}

In \cref{fig:common_corruptions}, we report the mean deviation of all tested models across the different data distributions averaged over different corruptions and samples, normalized by the maximum deviation achieved for each corruption type. We compute the mean deviation across all corruption schemes, normalized by the maximum deviation achieved for each individual corruption scheme, so as to weigh them evenly.
Both specialized models display nearly identical robustness across all datasets, confirming prior research on robustness under different data distributions. Most VLMs, in contrast, show non-uniform robustness with more fragility on AgeDB and FG-Net and far higher robustness on our dataset of known celebrities. \textit{Qwen 3.5} is an outlier, showing lower robustness only for AgeDB and similar robustness across FG-Net, CelebA and VideoScrape. A reason for that could be that it recognizes more celebrities on AgeDB than the other datasets.

Interestingly, robustness is similar on three different benchmark datasets (AgeDB, FG-Net, VideoScrape) for all VLMs except for \textit{Gemma~3}. Since VideoScrape is guaranteed to be excluded from the training sets, we can deduce that the benchmark datasets have been excluded from the training corpora of the studied VLMs as well. Lastly, the increased robustness on CelebA indicates that the impact of the identity shortcut is less prone to change under image corruptions than normal predictions. This implies that the memorization effect persists even when noise of various means is applied. It is worth noting that this finding holds even under various modifications of the inference prompt that we explored in preliminary experiments across the studied vision language models. However, we did not observe a significant difference for the identity shortcut, so we conclude that prompt design is not a viable explanation for the performance discrepancy.

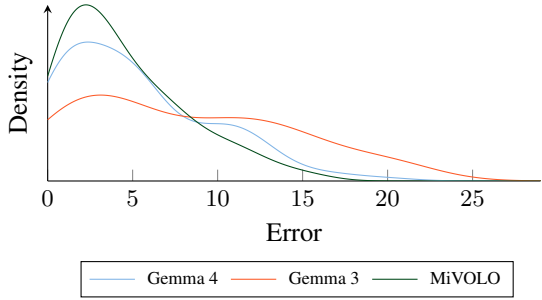
\begin{wrapfigure}{R}{0.5\columnwidth}
    % \addtocounter{figure}{1}
    \centering
    \begin{tikzpicture}
\pgfplotsset{every tick label/.append style={font=\small}}
\begin{axis}[
    xlabel={Error},
    ylabel={Density},
    axis x line*=bottom,
    axis y line*=left,
    ymin=0.0,
    xmin=0,
    xmax=29,
    width=0.58\columnwidth,
    height=0.28\columnwidth,
    ymajorgrids,
    legend cell align={left},
    axis y line=left,
    ytick=\empty,
    legend style={
        nodes={font=\footnotesize,scale=0.85,transform shape},
        legend columns=4,column sep=0.1cm
    },
    legend cell align={left},
    legend pos={north west},
    legend to name={errorDistLegend},
]
    \addplot [mark=none, poscolor] table [x index=0,y index=1] {data/error_dist/google_gemma-4-31B-it.dat};
    \addlegendentry{Gemma 4}
    \addplot [mark=none, negcolor] table [x index=0,y index=1] {data/error_dist/google_gemma-3-4b-it.dat};
    \addlegendentry{Gemma 3}
    \addplot [mark=none, neucolor] table [x index=0,y index=1] {data/error_dist/mivolo.dat};
    \addlegendentry{MiVOLO}
    
\end{axis}
    \node at (3.3, -1.3) {\pgfplotslegendfromname{errorDistLegend}};
\end{tikzpicture}
    \caption{Distribution of mean absolute error of selected models on FG-Net. While MiVOLO's error distribution has a single pronounced peak, the error distribution of the VLMs is bimodal.}
    % \addtocounter{figure}{-2}
    \label{fig:error_density}
    \vspace{-1.1cm}
\end{wrapfigure}

Besides the analysis presented above, we report the disaggregated results together with the adversarial robustness evaluation in Appendix~\ref{app:add-corrs}. In summary, the discovered trends present across the majority of corruptions for most of the models. Regarding adversarial robustness, we observe a similar discrepancy between CelebA and other datasets, though it is worth to mention how starkly non-robust all studied models are (all models are reliably attacked with \(\varepsilon < 1/255\)).

\subsection{Impact on Unknown Identities}\label{sec:meth:non-celeb}
As we show in Sections \ref{sec:meth:celeb} and \ref{sec:robustness}, there is a significant difference in model behavior for images with known and unknown people for vision-language models. However, even within the unknown people distribution, there is a discrepancy between traditional methods such as MiVOLO and VLMs. In Figure \ref{fig:error_density}, we show the model's error distribution on FG-Net, a dataset with a guaranteed absence of celebrities. The error distribution of MiVOLO behaves as expected, while the distribution for both \textit{Gemma~3} and \textit{Gemma~4} has a pronounced bimodal structure, which indicates that different behavior is at play even for unknown people in some cases. We hypothesize that models try to guess their identity and hence produce grossly incorrect predictions.

\section{Task Activation Steering}
\label{sec:meth:steering}

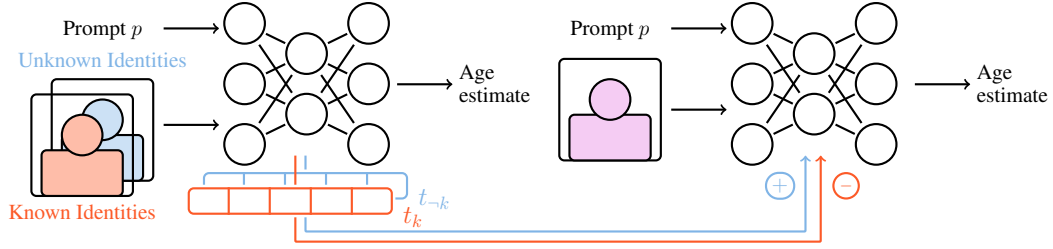
\begin{figure}
    \centering
    \resizebox{\linewidth}{!}{
\begin{tikzpicture}[
    node distance = 3.5mm and 4mm,
    font=\small,
    arr/.style = {->,very thick,
                  shorten >= 0.05mm,
                  shorten <= 0.05mm,
                  rounded corners=0.2mm},
    conn/.style = {very thick},
    box/.style = {rectangle, draw, very thick,
                 minimum height=8mm, minimum width=5mm,
                 rounded corners=1mm},
    circle/.style = {rectangle, draw, very thick,
                 minimum height=8mm, minimum width=8mm,
                 fill=white,rounded corners=4mm},
]

% --- Provided building blocks ---
\def\drawNN#1#2{
    \begin{scope}[shift={(#1,#2)}]
        \node (l1n1) [circle] {};
        \node (l1n2) [circle, below=of l1n1] {};
        \node (l1n3) [circle, below=of l1n2] {};
        \node (l2n1) [circle, right=0.8cm of $(l1n1)!0.5!(l1n2)$] {};
        \node (l2n2) [circle, below=of l2n1] {};
        \node (l3n1) [circle, right=1.6cm of l1n1] {};
        \node (l3n2) [circle, below=of l3n1] {};
        \node (l3n3) [circle, below=of l3n2] {};
        \draw[conn] (l1n1) -- (l2n1);
        \draw[conn] (l1n1) -- (l2n2);
        \draw[conn] (l1n2) -- (l2n1);
        \draw[conn] (l1n2) -- (l2n2);
        \draw[conn] (l1n3) -- (l2n1);
        \draw[conn] (l1n3) -- (l2n2);
        \draw[conn] (l2n1) -- (l3n1);
        \draw[conn] (l2n1) -- (l3n2);
        \draw[conn] (l2n1) -- (l3n3);
        \draw[conn] (l2n2) -- (l3n1);
        \draw[conn] (l2n2) -- (l3n2);
        \draw[conn] (l2n2) -- (l3n3);
    \end{scope}
}

\def\drawVector#1#2#3{
    \begin{scope}[shift={(#1,#2)}]
        \node (box) [box,minimum width=41mm,minimum height=6mm,white,fill=white] 
            at (0,0) {};
        \node (box) [box,minimum width=40mm,minimum height=5mm,#3,fill=white] 
            at (0,0) {};
        \draw[conn,#3] ($(box.north west)!0.2!(box.north east)$) -- ($(box.south west)!0.2!(box.south east)$);
        \draw[conn,#3] ($(box.north west)!0.4!(box.north east)$) -- ($(box.south west)!0.4!(box.south east)$);
        \draw[conn,#3] ($(box.north west)!0.6!(box.north east)$) -- ($(box.south west)!0.6!(box.south east)$);
        \draw[conn,#3] ($(box.north west)!0.8!(box.north east)$) -- ($(box.south west)!0.8!(box.south east)$);
    \end{scope}
}

\def\drawPerson#1#2#3{
    \begin{scope}[shift={(#1,#2)}]
        \node [box,minimum width=20mm,minimum height=20mm] {};
        \node [box,minimum height=9mm,minimum width=16mm,
               yshift=-5.5mm,fill=#3!40] {};
        \node [circle,minimum height=8mm,minimum width=8mm,rounded corners=4mm,
               yshift=2mm,fill=#3!40] {};
    \end{scope}
}

% --- Cross arrows (transfer) ---
\draw[poscolor,arr] (4.2,2.2) -- (4.2,0.7) -- (14.05,0.7) -- (14.05,2.2);
\drawVector{4.2}{1.6}{poscolor}
\draw[negcolor,arr] (4.0,2.2) -- (4.0,0.5) -- (14.35,0.5) -- (14.35,2.2);
\drawVector{3.9}{1.3}{negcolor}
% +/- signs
\node[circle,negcolor,fill=white,minimum width=5mm,minimum height=5mm,rounded corners=2.5mm] at (14.85,1.6) {\(-\)};
\node[circle,poscolor,fill=white,minimum width=5mm,minimum height=5mm,rounded corners=2.5mm] at (13.55,1.6) {\large \(+\)};

% --- Left: Task Vector computation ---
\begin{scope}
    \drawPerson{0.2}{2.7}{poscolor}
    \drawPerson{-0.2}{2.4}{negcolor}
    \node[poscolor,anchor=north] at (0.2,4.35) {\large Unknown Identities};
    \node[negcolor,anchor=north] at (-0.2,1.35) {\large Known Identities};
    \node[anchor=south] at (0.2,4.35) {\large Prompt \(p\)};

    \drawNN{3}{4.8}
    \node[anchor=north west,poscolor] at (6.3,1.65) {\Large \(t_{\neg k}\)};
    \node[anchor=north west,negcolor] at (6,1.3) {\Large \(t_k\)};

    \draw[arr] (1.4,4.7) -- (2.5,4.7);
    \draw[arr] (1.4,2.8) -- (2.5,2.8);
    \draw[arr] (6,3.6) -- (7.1,3.6);
    % \node[anchor=west] at (7.1,3.6) {\large Age estimate};
    \node[anchor=west] at (7.1,3.8) {\large Age};
    \node[anchor=west] at (7.1,3.4) {\large estimate};
\end{scope}

% --- Right: Inference ---
\begin{scope}
    \drawPerson{10.2}{3.1}{color3}
    \node[anchor=south] at (10.2,4.35) {\large Prompt \(p\)};

    \drawNN{13}{4.8}

    \draw[arr] (11.4,4.7) -- (12.5,4.7);
    \draw[arr] (11.4,3.1) -- (12.5,3.1);
    \draw[arr] (16.2,3.6) -- (17.3,3.6);
    % \node[anchor=west] at (18.3,3.6) {\large Age};
    \node[anchor=west] at (17.3,3.8) {\large Age};
    \node[anchor=west] at (17.3,3.4) {\large estimate};
\end{scope}

\end{tikzpicture}%
}
    \caption{Overview of our task activation steering method. The VLM computes a task vector from the image and prompt, based on which the age is estimated. We modify this task vector by moving it towards the distribution of unknown identities, effectively disabling the identity shortcut.}
    \label{fig:overview}
\end{figure}

To further analyze the mechanism behind the identity shortcut, we utilize task vectors, first introduced by \citet{taskVectors}. The concept of task vectors is based on the idea of decomposing a language model~\(f\), which receives a task description \(p\) and an input sample \(x\) into two functions applied in sequence. The first function, \(t(\cdot)\) is responsible for extracting a task vector from the task description which is then used to produce the final prediction, so that \(f(x, p) = a(x, t(p)).\)
Based on our findings, we extend this definition to task vectors also dependent on the input, i.e. we decompose \(f\) by \begin{equation}
    f(x) = a(x, t(x, p)).
\end{equation}

This formulation allows us to describe the identity shortcut as a different task vector being produced depending on whether the identity is known or unknown. Thus, we define the task vector when the identity is known as \(t_k\) and when it is unknown as \(t_{\neg k}\), resulting in \begin{equation}
    f(x \mid k) = a(x, t_k), \quad f(x\mid \neg k) = a(x, t_{\neg k}).
\end{equation}

From an implementation point of view, we base our task vector computation on the findings by \citet{taskVectors}, which say that the shallow activations at the last token position of the prompt are responsible for encoding the task. While \citet{taskVectors} use activations from a single layer, we instead take all activations of the first half of the model at this last token position as the task vector to get a robust representation and avoid the search over layer indices. Then, we can compute the task vector for known and unknown identities by
\begin{equation}
    t_k = \mathop{\mathbb{E}}_{x\sim \mathbb{K}_f} t(x, p), \quad t_{\neg k} = \mathop{\mathbb{E}}_{x\sim \mathbb{X}\setminus\mathbb{K}_f} t(x, p).
\end{equation}

\subsection{Task Vector Distribution}
Utilizing the introduced task vectors, we can now investigate the cause of the unusual error distribution for non-celebrity images.
We measure the number of samples for which the predicted task vector \(t(x)\) is closer to the identity shortcut task vector \(t_k\) than the task vector for unknown identities \(t_{\neg k}\). To express it as a single-dimensional value, we define the difference in distance to the task vectors \begin{equation}
    \Delta_k(t) = \|t - t_{\neg k}\| - \|t - t_{k}\|.
\end{equation}
For task vectors closer to the identity shortcut we get \(\Delta_k(t) < 0\) and for those further from the identity shortcut, we have \(\Delta_k(t) > 0\). To account for the spread of task vectors, we additionally define the distributions \(\mathbb{T}_k\) and \(\mathbb{T}_{\neg k}\) for task vectors on known and unknown identities of AgeDB, respectively.

\begin{figure}[bt]
    \centering
    \begin{subfigure}[t]{0.48\columnwidth}
        \centering
        \begin{tikzpicture}
\pgfplotsset{every tick label/.append style={font=\small}}
\begin{axis}[
    xlabel={\(\Delta_k\)},
    ylabel={Density},
    axis x line*=bottom,
    axis y line*=left,
    ymin=0.0,
    xmin=-27,
    xmax=29,
    width=0.99\columnwidth,
    height=0.48\columnwidth,
    ymajorgrids,
    legend cell align={left},
    axis y line=left,
    ytick=\empty,
    legend style={
        nodes={font=\footnotesize,scale=0.85,transform shape},
        legend columns=4,column sep=0.1cm
    },
    legend cell align={left},
    legend pos={north west},
    legend to name={misidentificationLegend},
]
    \addplot [mark=none, poscolor] table [x index=0,y index=2] {data/misidentification/google_gemma-4-31B-it.dat};
    \addlegendentry{\(\mathbb{T}_k\)}
    \addplot [mark=none, negcolor] table [x index=0,y index=3] {data/misidentification/google_gemma-4-31B-it.dat};
    \addlegendentry{\(\mathbb{T}_{\neg k}\)}
    \addplot [mark=none, neucolor] table [x index=0,y index=6] {data/misidentification/google_gemma-4-31B-it.dat};
    \addlegendentry{FG-Net}
    \addplot [mark=none, color3] table [x index=0,y index=7] {data/misidentification/google_gemma-4-31B-it.dat};
    \addlegendentry{VideoScrape}
    
\end{axis}
    \node at (2.7, -1.3) {\pgfplotslegendfromname{misidentificationLegend}};
\end{tikzpicture}%
        \caption{Density of different task vectors for Gemma 4 along the axis \(t_k-t_{\neg k}\) of different datasets. FG-Net and VideoScrape are closer to unknown identities but still have some overlap with known identities.}
        \label{fig:misidentification_density}
    \end{subfigure}~~
    \begin{subfigure}[t]{0.48\columnwidth}
        \centering
        \begin{tikzpicture}
\pgfplotsset{every tick label/.append style={font=\small}}
\begin{axis}[
    xlabel={Error},
    ylabel={Density},
    axis x line*=bottom,
    axis y line*=left,
    ymin=0.0,
    xmin=0,
    xmax=29,
    width=0.99\columnwidth,
    height=0.48\columnwidth,
    ymajorgrids,
    legend cell align={left},
    axis y line=left,
    ytick=\empty,
    legend style={
        nodes={font=\footnotesize,scale=0.85,transform shape},
        legend columns=4,column sep=0.1cm
    },
    legend cell align={left},
    legend pos={north west},
    legend to name={errorDistFixLegend},
]
    \addplot [mark=none, poscolor] table [x index=0,y index=1] {data/error_dist/google_gemma-4-31B-it_fix.dat};
    \addlegendentry{Gemma 4}
    \addplot [mark=none, negcolor] table [x index=0,y index=1] {data/error_dist/google_gemma-3-4b-it_fix.dat};
    \addlegendentry{Gemma 3}
    \addplot [mark=none, dashed, poscolor] table [x index=0,y index=1] {data/error_dist/google_gemma-4-31B-it.dat};
    \addplot [mark=none, negcolor, dashed] table [x index=0,y index=1] {data/error_dist/google_gemma-3-4b-it.dat};
    % \addplot [mark=none, neucolor] table [x index=0,y index=1] {data/error_dist/mivolo.dat};
    % \addlegendentry{MiVOLO}
    
\end{axis}
    \node at (2.6, -1.3) {\pgfplotslegendfromname{errorDistFixLegend}};
\end{tikzpicture}%
        \caption{Error distribution of selected models on FG-Net with (solid line) and without (dashed) activation steering. The steered model error distribution becomes closer to normal.}
        \label{fig:error_density_fix}
    \end{subfigure}
    \caption{Effects of the identity shortcut on photos containing unknown identities.}
\end{figure}
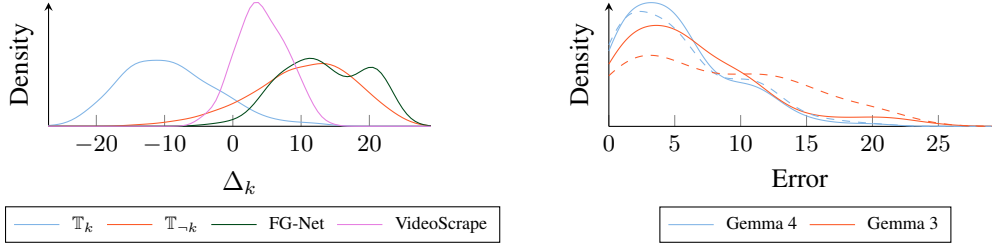

\cref{fig:misidentification_density} displays example distributions of task vectors computed from \textit{Gemma 4} on known identities from AgeDB, unknown identities and two datasets consisting of unknown identities, FG-Net and VideoScrape. We see that there is a clear separation between known and unknown identities of AgeDB. The FG-Net distribution is close to the unknown identities distribution, while VideoScrape, is still closer to the unknown identities distribution, but situated further in between both. This indicates that \textit{Gemma 4} does take the identity shortcut for some samples, even though the identity cannot be known by the model and must thus be misidentified. To show how many samples activate the identity shortcut, we test \(\mathbb{P}(t \in \mathbb{T}_{\neg k}) < 0.1 \land \mathbb{P}(t \in \mathbb{T}_k) > 0.1\). The resulting ratios are shown in \cref{tab:misidentification}.

\begin{table}[tbh]
    \centering
    \caption{Ratio of samples for which the task vector indicates usage of the identity shortcut.}
    \label{tab:misidentification}
    \begin{tabular}{l ccccc}
    \toprule
    Dataset & LLaVA 1.5 & QwenVL 2.5 & Gemma 3 & Qwen 3.5 & Gemma 4 \\
    \midrule
    FG-Net &
    6.98\% & 0.58\% & 0.00\% & \phantom{1}1.16\% & \phantom{1}1.16\% \\
    VideoScrape &
    6.82\% & 0.00\% & 0.91\% & 17.73\% & 10.91\% \\
    \bottomrule
\end{tabular}
\end{table}

We see that all models have samples on which the identity shortcut is prominent, but \textit{QwenVL~2.5} and \textit{Gemma~3} have particularly small ratios. While there is no clear trend along the models, \textit{Qwen~3.5} and \textit{Gemma~4} have the highest ratios with \textit{Qwen~3.5} activating the identity shortcut for nearly 18\% of the samples in the VideoScrape dataset, additionally reinforcing our suggestion in Section \ref{sec:meth:celeb}, that \textit{Qwen~3.5} is especially prone to the shortcut even on datasets without celebrities present.

\subsection{Disabling the Identity Shortcut}
As the identity shortcut has a detrimental impact on model performance, it is beneficial to remove it. One possible way to do so, would be to remove the identity knowledge from the model during the training process or via machine unlearning techniques. This approach, however, is either computationally heavy to implement (full retraining), or introduces additional loss of performance (machine unlearning). Among the inference-time techniques, prompt-engineering could be used. However as we discuss in Section \ref{sec:robustness}, our attempts to change the prompt did not affect the model prediction, even if we prompted the model to ignore its knowledge of the person or focus on the visual appearance in different ways.
Instead, we propose to keep the model on the correct task, via an inference-time activation-level intervention. The high-level process is explained in \cref{fig:overview}. We use the task vectors \(t_k\) and \(t_{\neg k}\) to intervene in the hidden states and move the model from the identity shortcut distribution towards the distribution of task vectors with unknown identity. More concretely, we approximate age estimation with unknown identity via \begin{equation}
    f(x\mid \neg k) \approx a(x, t(x, p) + \alpha \cdot (t_{\neg k} - t_k)),
\end{equation}
where \(\alpha\) is a hyperparameter. We choose \(\alpha=3\) through manual tuning on a set of manually collected validation images of celebrities and non-celebrities that are not included in any of the benchmark datasets. We apply this activation steering to the VLMs during age estimation and compute the mean absolute error on AgeDB and FG-Net. The results are displayed in \cref{tab:steering}.

\begin{table}[htb]
    \centering
    \caption{Mean absolute error of vision language models on AgeDB and FG-Net in the default mode and with activation steering to fix shortcut mispredictions.}
    \begin{tabular}{l l ccccc}
    \toprule
    Dataset & Variant & LLaVA 1.5 & QwenVL 2.5 & Gemma 3 & Qwen 3.5 & Gemma 4 \\
    \midrule
    \multirow{2}{*}{AgeDB}
    & Default
    & 9.21 $\pm$ 0.06 & 8.94 $\pm$ 0.06 & 8.48 $\pm$ 0.05 & 13.52 $\pm$ 0.09 & \textbf{6.50} $\pm$ 0.04 \\
    & Steered
    & \textbf{8.93} $\pm$ 0.06 & \textbf{7.78} $\pm$ 0.05 & \textbf{8.01} $\pm$ 0.05 & \textbf{10.72} $\pm$ 0.07 & 6.73 $\pm$ 0.04 \\[1.5ex]
    
    \multirow{2}{*}{FG-Net}
    & Default
    & 7.06 $\pm$ 0.50 & 9.03 $\pm$ 0.58 & 8.36 $\pm$ 0.57 & 8.80 $\pm$ 0.54 & 5.23 $\pm$ 0.38 \\
    & Steered
    & 7.13 $\pm$ 0.49 & \textbf{7.02} $\pm$ 0.50 & \textbf{6.24} $\pm$ 0.45 & \textbf{7.17} $\pm$ 0.48 & \textbf{5.08} $\pm$ 0.35 \\
    \bottomrule
\end{tabular}
    \label{tab:steering}
\end{table}

The resulting errors show that steering the activations towards \(t_{\neg k}\) and away from \(t_k\) does indeed help performance in almost all cases. The mean improvement across all setups is \(10.8\%\), showing the measurable practical impact of the proposed method. \textit{Gemma 3} sees the biggest improvement with a \(25.36\%\) reduction of error on FG-Net. This indicates that the identity shortcut causes mispredictions, both for celebrities on AgeDB and non-celebrities on FG-Net. Interestingly, we see no clear correlation between the ratio of samples for which the task vector indicates usage of the identity shortcut and the performance uplift. We suspect that older models, such as LLaVa~1.5 are not capable enough to perform well on age estimation and therefore removing the identity shortcut only reverts them back to their poor baseline performance. Lastly, Qwen~3.5 shows the nearly biggest improvement, providing additional evidence of our previous findings regarding this model.

Further analysis on AgeDB shows that improvements for correctly identified people are higher than for misidentified samples, confirming that the identity shortcut's impact on celebrities is higher than on non-celebrities. Nevertheless, improvements for all models except LLava 1.5 on FG-Net show that the effect on non-celebrities is prominent. It is worth mentioning that activation steering has little impact on the inference time and therefore has a negligible overhead for real-world deployment. In addition, we show the resulting error distribution when applying activation steering in \cref{fig:error_density_fix}. It is apparent that the bimodal nature of the error distribution is dampened. Therefore, we conclude that the identity shortcut is indeed the reason for its appearance in the first place.

\section{Conclusion}
\label{sec:conclusion}
It is hard to overlook the progress in zero-shot computer vision enabled by vision-language models. Age estimation is a particularly revealing case: associations acquired during large-scale pretraining align well with the task, and zero-shot VLMs now define the state of the art.
However, as we show in this paper, some of these associations play an insidious role and lead to erroneous results.
The identity shortcut we expose is one such phenomenon, and we expect it is only the first of many failure modes to surface within the internal mechanisms of VLMs. Our activation-steering method addresses the shortcut, offering a path toward correcting VLM failure modes more broadly.

From a broader perspective, our results point to an increasing dilemma in AI research. As learning models grow in size and capability, we are departing from task-specific learning, where the inference process is dictated by the employed architecture. VLMs and other large models become so flexible, that they are capable of solving the same task through different means. This is an unmatched advantage technically; at the same time, it can undermine our ability to measure genuine progress if models solve tasks in unexpected or even incorrect ways.

\paragraph{Societal Impact.} Our robustness analysis further shows that current VLM-based age estimators fail under low-magnitude perturbations, with error rates unacceptable for security-critical systems. Given that automated age estimation is already being deployed under recent legislation governing minors' access to online services, the resulting gap between benchmark performance and real-world reliability creates a deployment risk that cannot be ignored. We therefore echo the community's concerns about employing age estimators in this setting.

\paragraph{Limitations and Future Work.} One limitation of our method, is its reliance on open-weight models. Even though the technique is applied during inference time, it still requires access to the activations, limiting its use for the best-performing closed-weights VLMs. Another limitation of our work is that we study the identity shortcut only on one task---age estimation. Even though this setup is important for real-world deployments, we hypothesize that these identity shortcuts generalize to other attribute inference tasks like gender, height and weight. Rigorous validation across a wider array of models, datasets, and demographic distributions remains a critical direction for future research.%we conjecture that phenomenon itself is present in other tasks with inferring different attributes from images (gender, height, weight, etc), and extending our study to these tasks is a direction for the future work.

% \begin{ack}
% This work was supported by the German Federal Ministry of Research, Technology and Space under the grant ...
% \end{ack}

% \section*{References}
\bibliographystyle{abbrvnat}
\bibliography{references}
\clearpage

\appendix
\section{Experimental Details}\label{sec:appendix:details}
To aid reproducibility and foster further research in this area, we aim to release as many details about our experiments as possible. Any details not found in this section may be inferred from the provided code. We make an anonymized version available for the time of the reviews under \href{https://anonymous.4open.science/r/identity-shortcut-4D53}{anonymous.4open.science/r/identity-shortcut-4D53}.

\subsection{Generation Settings}

To query the VLMs and ensure reproducible, comparable results between them, we always utilized the same prompt: \textbf{``Estimate the age of the person in this photograph. Respond with ONLY a single integer representing their age in years. Do not include any other text, explanation, or units''}. Additionally, we set the temperature to 0 and the maximum number of generated tokens to 10. We then parse the estimated age as an integer from the response. We report the precisions of the models and their licenses in \cref{tab:licenses_models}. For activation steering, we use \(\alpha=3\) across all models.

\begin{table}[tbh]
    \centering
    \caption{Models, their precision, and licenses.}
    \begin{tabular}{lcc}
    \toprule
    \makecell[c]{Model} & \makecell[c]{Precision} & \makecell[c]{License} \\
    \midrule
    ResNet-18      & bfloat16 & BSD-3-Clause\\
    MiVOLO         & bfloat16 & Apache-2.0\\
    \midrule
    LLaVa 1.5 7B   & bfloat16 & Apache-2.0\\
    QwenVL 2.5 7B  & bfloat16 & Apache-2.0 \\
    Gemma 3 4B     & bfloat16 & \href{https://ai.google.dev/gemma/terms}{Gemma License} \\
    Qwen 3.5 9B    & quantized 4 bit & Apache-2.0 \\
    Gemma 4 31B    & quantized 4 bit & Apache-2.0 \\
    Gemini 3 Flash & n/a & \href{https://ai.google.dev/gemini-api/terms}{Gemini Terms of service}\\
    \bottomrule
    \end{tabular}
    \label{tab:licenses_models}
\end{table}

\subsection{Identifying Identity Knowledge}
We identify when a VLM knows the identity of a person in two different ways, depending on the availability of labels. For AgeDB, labels with the person's name exist, so we simply prompt the model to guess the identity of the depicted person and accept answers with an edit distance of less than 5. This allows for some leeway in the names such as accents on characters being included or omitted in the ground truth and model answer while keeping the number of false positives to a minimum. The prompt used for this task is \textbf{``Do you know this person? If yes, provide their name only. If not, say 'Unknown''}.

On CelebA, we do not have the correct labeled identities, so we instead check the answers by cross-checking the answer through a separate prompt in a separate context. For this, we use the following prompt: \textbf{``Is this person 'NAME'? Answer with yes or no only.''}, where ``NAME'' is the answer provided by the previous prompt. Manual checks of ten randomly chosen samples confirm that the person is indeed identified correctly.

\subsection{Robustness Settings}
In order to evaluate the performance of the VLMs under common noise corruptions, we manipulated the images from the utilized datasets with the corruptions listed in \cref{tab:corruptions}. We analyzed several severity levels, namely severity levels $\in\{0.25, 0.5, 0.75, 0.99\}$. Depending on these levels, we calculated parameters for the different corruptions. We report the values for the parameters or parameter pairs for each of the four severity levels. As displayed in \cref{fig:common_corruptions}, we average over all these corruption functions and severity levels, as well as over the applied adversarial perturbations.

\begin{table}[ht]
\centering
\scriptsize
\caption{Common noise corruptions and the hyperparameters depending on our four severity levels.}
\begin{tabular}{llp{0.643\linewidth}}
\toprule
\makecell[c]{Corruption} & \makecell[c]{Parameter} & \makecell[c]{Values given severity}\\
\midrule
Gaussian Noise & $\sigma$ & $\{0.13, 0.22, 0.31, 0.40\}$ \\

Shot Noise & $c$ & $\{17.25, 31.50, 45.75, 59.43\}$ \\

Impulse Noise & $c$ & $\{0.09, 0.15, 0.21, 0.27\}$ \\

Speckle Noise & $c$ & $\{0.26, 0.38, 0.49, 0.60\}$ \\

\midrule

Defocus Blur & (radius, alias blur) & $\left\{\begin{array}{cc} 4.75, & 0.20 \\
6.50, & 0.30 \\
8.25, & 0.40 \\
9.93, & 0.50 \end{array}\right\}$ \\

Glass Blur & ($\sigma$, $\max\delta$, iterations) & $\left\{\begin{array}{ccc} 0.90, & 1, & 2 \\
1.10, & 2, & 2 \\
1.30, & 3, & 2 \\
1.49, & 3, & 2 \end{array}\right\}$ \\

Motion Blur & (radius, $\sigma$) & $\left\{\begin{array}{cc} 12.50, & 6.00 \\
15.00, & 9.00 \\
17.50, & 12.00 \\
19.90, & 14.88 \end{array}\right\}$ \\

Zoom Blur & zoom factor & $\left\{\begin{array}{ccccccccc} 1.00, & 1.01, & 1.03, & 1.04, & 1.06, & 1.07, & 1.09, & 1.10 & \\
1.00, & 1.02, & 1.04, & 1.06, & 1.08, & 1.10, & 1.12, & 1.14, & 1.16 \\
1.00, & 1.02, & 1.05, & 1.07, & 1.10, &, 1.12, & 1.15, & 1.17, & 1.20 \\
1.00, & 1.03, & 1.06, & 1.09, & 1.12, & 1.15, & 1.18, & 1.21, & 1.24 \end{array}\right\}$ \\

Gaussian Blur & $\sigma$ & $\{2.25, 3.50, 4.75, 5.95\}$ \\

\midrule

Snow & $(c_1, \dots, c_7)$ & $\left\{\begin{array}{ccccccc} 0.21, & 0.30, & 3.38, & 0.59, & 10.50, & 5.00, & 0.94 \\
0.33, & 0.30, & 3.75, & 0.68, & 11.00, & 6.00, & 1.08 \\
0.44, & 0.30, & 4.12, & 0.76, & 11.50, & 7.00, & 1.21 \\
0.55, & 0.30, & 4.48, & 0.85, & 11.98, & 7.96, & 1.34 \end{array}\right\}$ \\

Frost & $(c_1, c_2)$ & $\left\{\begin{array}{cc} 0.90, & 0.49 \\
0.80, & 0.57 \\
0.70, & 0.66 \\
0.60, & 0.75 \end{array}\right\}$ \\

Fog & $(c_1, c_2)$ & $\left\{\begin{array}{cc} 1.88, & 1.85 \\
2.25, & 1.70 \\
2.62, & 1.55, \\
2.98, & 1.41 \end{array}\right\}$ \\

Spatter & $(c_1, \dots, c_6)$ & $\left\{\begin{array}{cccccc} 0.66, & 0.33, & 1.75, & 0.68, & 0.82, & 0.00 \\
0.66, & 0.35, & 2.50, & 0.67, & 1.05, & 0.00 \\
0.67, & 0.38, & 3.25, & 0.67, & 1.27, & 0.00 \\
0.67, & 0.40, & 3.97, & 0.66, & 1.49, & 0.00 \end{array}\right\}$ \\

\midrule

Brightness & $c$ & $\{0.20, 0.30, 0.40, 0.50\}$ \\

Contrast & $c$ & $\{0.31, 0.23, 0.14, 0.05\}$ \\

Saturate & $(c_1, c_2)$ & $\left\{\begin{array}{cc} 5.22, & 0.05 \\
10.15, & 0.10 \\
15.07, & 0.15 \\
19.80, & 0.20 \end{array}\right\}$ \\

\midrule

Elastic & $(c_1, c_2, c_3)$ & $\left\{\begin{array}{ccc} 0.80, & 4.80, & 2.16 \\
1.60, & 3.20, & 1.76 \\
2.40, & 1.60, & 1.36 \\
3.17, & 0.06, & 0.98 \end{array}\right\}$ \\

Pixelate & $c$ & $\{0.46, 0.33, 0.19, 0.06\}$ \\

JPEG & quality & $\{20, 16, 12, 7\}$ \\

\midrule

Adversarial & \(\epsilon\) & $\{0.1/255, 0.3/255, 1/255, 8/255\}$ \\

\bottomrule
\end{tabular}
\label{tab:corruptions}
\end{table}

\subsection{Datasets}\label{sec:appendix:datasets}
We acquire AgeDB~\citep{moschoglou2017agedb}. FG-Net~\citep{Lanitis2002}, and CelebA~\citep{celebA} from \href{https://www.kaggle.com/datasets/}{Kaggle}. They are widely-used image datasets of celebrities and non-celebrities. AgeDB, in particular, consists of over 16\,000 images of 570 subjects labeled with age and gender. FG-Net is composed of 1\,002 images of 82 subjects. For most subjects, the dataset contains multiple photos taken at different stages of their lives. CelebA contains over 200\,000 images of a total of over 10\,000 individuals. These images are not annotated with a name, age, or gender.

\begin{table}[ht]
    \centering
    \caption{Datasets and their licenses.}
    \label{tab:licenses}
    \begin{tabular}{ll}
    \toprule
    \makecell[c]{Dataset} & \makecell[c]{License} \\
    \midrule
    AgeDB       & non-commercial research purposes only\\
    FG-Net      & MIT\\
    CelebA      & non-commercial research purposes only\\
    %AFAD        & academic research purposes only\\
    VideoScrape & \href{https://www.zerozerorobotics.com/terms}{Terms Of Use}, \href{https://www.zerozerorobotics.com/privacy}{Privacy Policy}\\
    \bottomrule
    \end{tabular}
\end{table}

% AFAD~\citep{afad2016},AFAD contains over 160,000 images annotated with age and gender, only including faces of humans of Asian decent.

\paragraph{VideoScrape}
To ensure that we evaluate our hypotheses on images of real human subjects which are not included in the training dataset of the analyzed VLMs, i.e. published after the knowledge cut-off date, we compose a dataset of 200 images acquired from publicly uploaded videos from autonomous drones. Owners of the HoverAir models from \href{https://www.zerozerorobotics.com/}{ZeroZero Robotics} can upload videos within the Hover app and share them with third parties via URLs. We use a Python script that semi-automatically browses the app for videos, copies, and saves the share link to a file. We then automatically go through all acquired links and check the metadata for the upload date and the actual video content URL. 

If the upload date is after the latest known VLM cutoff date, we download the video. We then center crop each frame to a 500 pixel wide square and detect faces within the cropped frame via a pretrained MTCNN~\citep{MTCNN2016} (MIT License). According to 7 (m) of the \href{https://www.zerozerorobotics.com/terms}{Terms Of Use} of ZeroZero Robotics, retrieving data of their websites or apps is prohibited only if it was ``...not made generally available to visitors to the Sites or to Registrants...''. If users actively decide to share their videos and photos captured with a HoverAir drone within the app, the content is publicly shared with all registered users to receive likes and comments. The code to retrieve the images is published within our code base. We are not storing any videos and we will delete the images of human subjects with the publication of this paper.

\paragraph{Subsampling}
To ensure a roughly equal distribution of ages and genders, we analyze our VideoScrape dataset using MiVOLOv2 and obtain the demographics as presented in \cref{tab:demo_hover}. We then analyze AgeDB and FG-Net and sample images randomly to match the VideoScrape demographics given enough images for a certain age bin are present in the dataset. FG-Net does not officially label their images with the ground-truth gender but the extended FG-Net Mask dataset~\citep{jimaging7100204} annotates the images accordingly.
\begin{table}[ht]
    \centering
    \caption{Demographics for the utilized datasets. We first analyze our VideoScrape dataset and then subsample images from AgeDB and FG-Net to match the same distribution as best as possible.}
    \begin{tabular}{lcccccccc}
\toprule
 & \multicolumn{8}{c}{Age Bins} \\
 & 0-2 & 3-6 & 7-12 & 13-20 & 21-32 & 33-43 & 44-53 & 54-100 \\
\midrule
\textbf{VideoScrape} & & & & & & & & \\
Male   & 0   & 1   & 5    & 1     & 25    & 55    & 38    & 41     \\
Female & 0   & 3   & 4    & 1     & 15    & 18    & 5     & 3     \\
\midrule
\textbf{AgeDB} & & & & & & & & \\
Male & 0 & 1 & 5 & 1 & 25 & 55 & 38 & 41 \\
Female & 0 & 3 & 4 & 1 & 15 & 18 & 5 & 3 \\
\midrule
\textbf{FG-Net} & & & & & & & & \\
Male & 0 & 1 & 5 & 1 & 25 & 38 & 19 & 10 \\
Female & 0 & 3 & 4 & 1 & 15 & 18 & 5 & 3 \\
% \midrule
% \textbf{AFAD} & & & & & & & & \\
% Male & 0 & 0 & 0 & 1 & 25 & 55 & 38 & 41 \\
% Female & 0 & 0 & 0 & 1 & 15 & 18 & 5 & 3 \\
\bottomrule
\end{tabular}
    \label{tab:demo_hover}
\end{table}

\subsection{Compute Resources}\label{sec:compute-resources}
For all experiments except those involving Gemini 3 Flash, we utilize a cluster of NVIDIA GPUs consisting of H100, A100, A40, H200 and RTX PRO 6000 Blackwell GPUs. We did not control which GPU was used for each experiment and therefore report aggregated GPU hours instead in \cref{tab:compute}. For Gemini 3 Flash, we use a total of \(109.3\) million input tokens and \(199.9\) thousand output tokens.
\begin{table}[htb]
    \centering
    \caption{Compute required for running the experiments.}
    \label{tab:compute}
    \begin{tabular}{lr}
        \toprule
        \makecell[c]{Experiment} & \makecell[c]{GPU Hours} \\
        \midrule
        Impact on Known Identities      & 4 \\
        Robustness across Distributions & 1\,499 \\
        Impact on Unknown Identities    & 2 \\
        Task Vector Distribution        & 2 \\
        Disabling the Identity Shortcut & 8 \\
        \bottomrule
    \end{tabular}
\end{table}
\FloatBarrier
\section{Additional Results}
\label{app:add-corrs}
\begin{figure}[htb]
    \centering
    \input{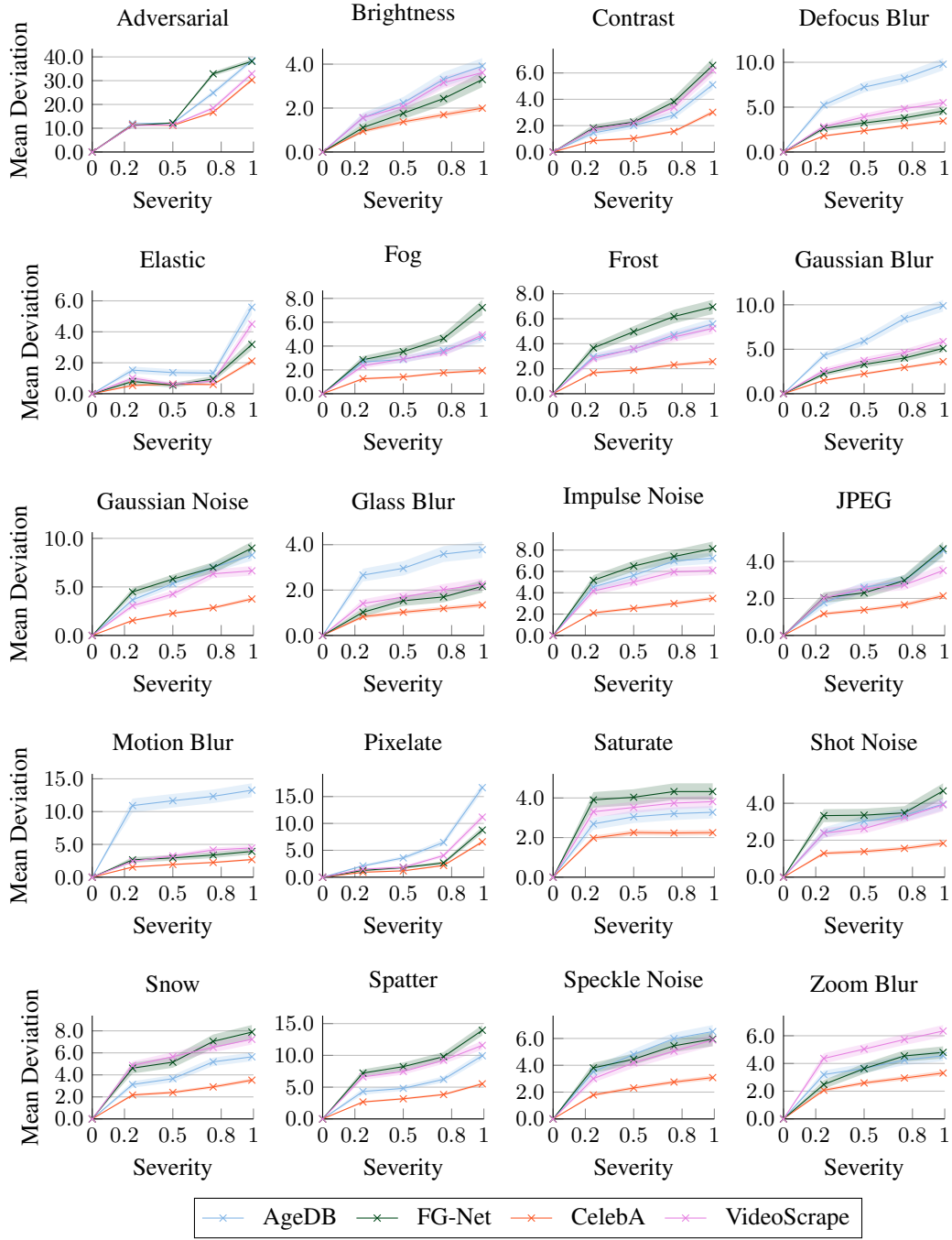}
    \caption{Detailed results for the deviations achieved by each corruption for LLaVa~1.5.}
    \label{fig:appendix:last}
\end{figure}

We provide additional detailed results across all models and corruption modes in Figures~\ref{fig:appendix:first}-\ref{fig:appendix:last}.

\begin{figure}[htb]
    \centering
    \input{figures/appendix/gemma3}
    \caption{Detailed results for the deviations achieved by each corruption for Gemma 3.}
    \label{fig:appendix:first}
\end{figure}

\begin{figure}[htb]
    \centering
    \input{figures/appendix/gemma4}
    \caption{Detailed results for the deviations achieved by each corruption for Gemma 4.}
\end{figure}

\begin{figure}[htb]
    \centering
    \input{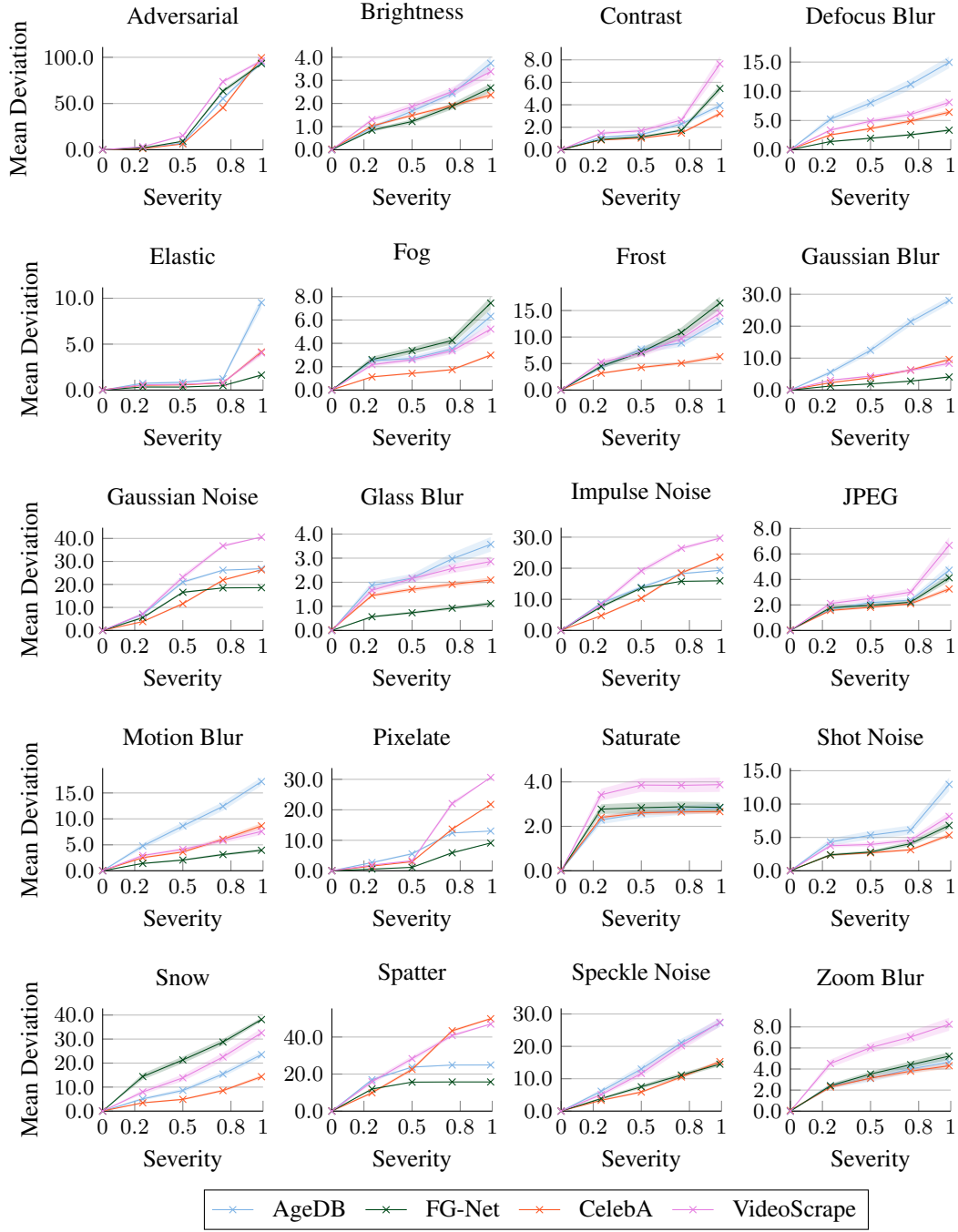}
    \caption{Detailed results for the deviations achieved by each corruption for MiVOLO.}
\end{figure}

\begin{figure}[htb]
    \centering
    \input{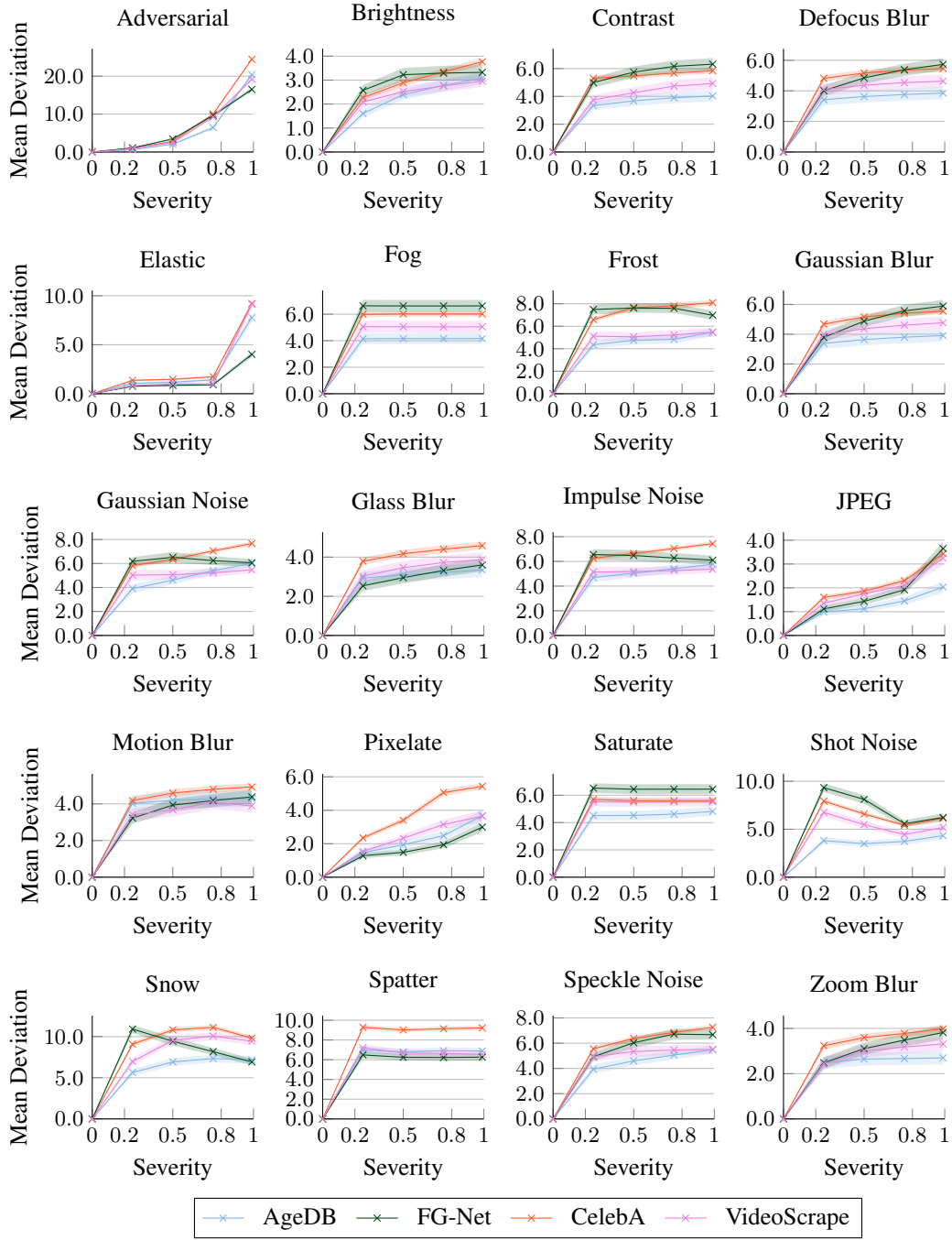}
    \caption{Detailed results for the deviations achieved by each corruption for the CNN.}
\end{figure}

\begin{figure}[htb]
    \centering
    \input{figures/appendix/qwen25}
    \caption{Detailed results for the deviations achieved by each corruption for QwenVL~2.5.}
\end{figure}

\begin{figure}[htb]
    \centering
    \input{figures/appendix/qwen35}
    \caption{Detailed results for the deviations achieved by each corruption for Qwen~3.5.}
\end{figure}

\begin{figure}[htb]
    \centering
    \input{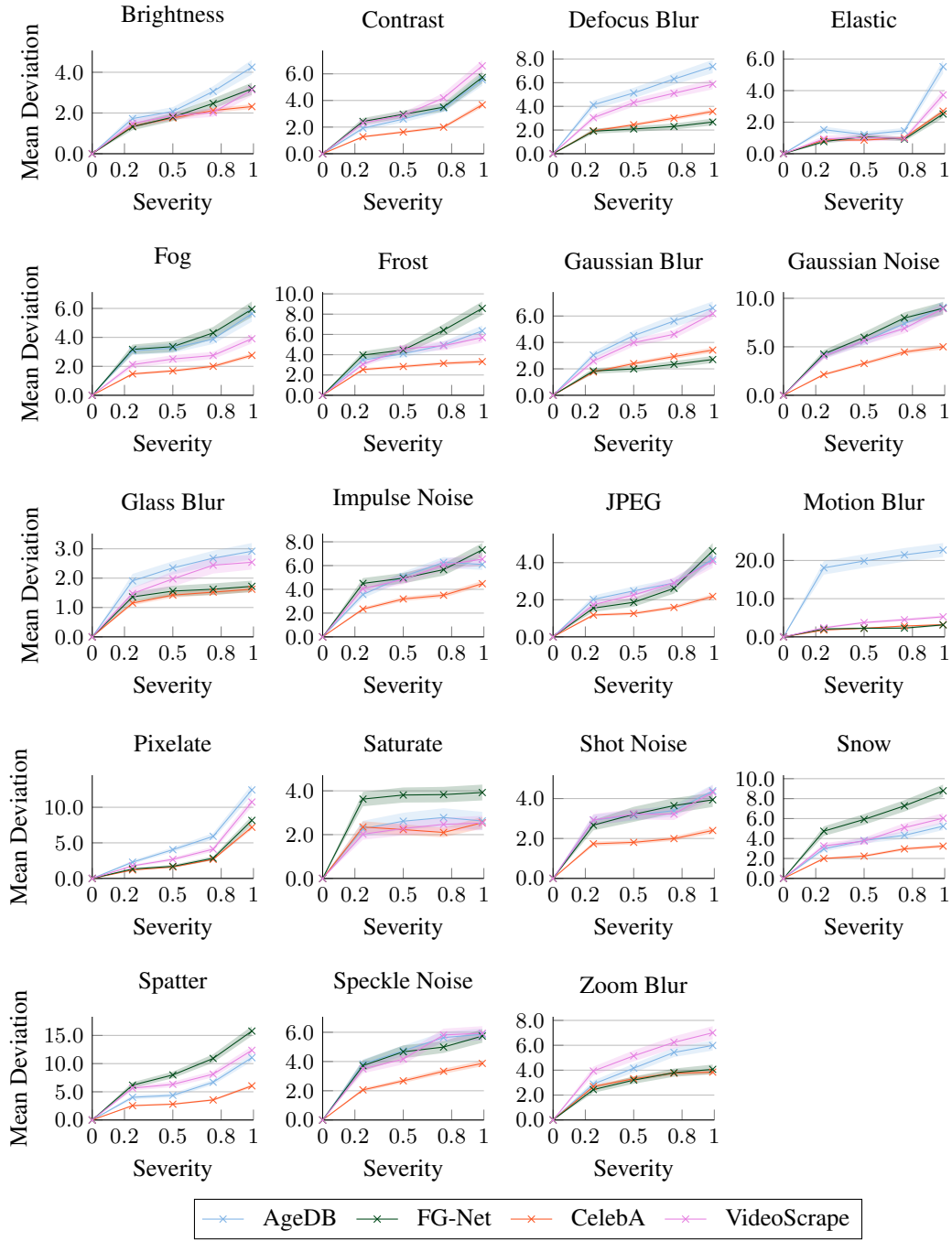}
    \caption{Detailed results for the deviations achieved by each corruption for Gemini 3 Flash.}
\end{figure}

% \clearpage
% \input{checklist.tex}

\end{document}